# MULTI-TAPE TWO-LEVEL MORPHOLOGY:
# A Case Study in Semitic Non-linear Morphology

George Anton Kiraz*

COMPUTER LABORATORY, UNIVERSITY OF CAMBRIDGE

(St John's College)

E-mail. `George.Kiraz@cl.cam.ac.uk`

July 26, 1994

## Abstract

This paper presents an implemented multi-tape two-level model capable of describing Semitic non-linear morphology. The computational framework behind the current work is motivated by [Kay 1987]; the formalism presented here is an extension to the formalism reported by [Pulman and Hepple 1993]. The objectives of the current work are: to stay as close as possible, in spirit, to standard two-level morphology, to stay close to the linguistic description of Semitic stems, and to present a model which can be used with ease by the Semitist. The paper illustrates that if finite-state transducers (FSTs) in a standard two-level morphology model are replaced with multi-tape auxiliary versions (AFSTs), one can account for Semitic root-and-pattern morphology using high level notation.

## 1 INTRODUCTION

This paper aims at presenting a computational morphology model which can handle the non-linear phenomenon of Semitic morphology. The approach presented here builds on two-level morphology [Koskenniemi 1983], extending it to achieve the desired objective. The contribution of this paper may be summarised as follows:

With regards to the two-level model, we extend this model by allowing it to have multiple tapes on the lexical level and retaining the one tape on the surface level; hence, 'multi-tape two-level morphology'. Feasible pairs in the standard two-level model become 'feasible tuple pairs' in our multi-tape model.

With regards to the formalism, we have chosen a two-level formalism and extended it to be able to write multi-tape two-level grammars which involve non-linear operations. To achieve this, we made all lexical expressions $n$-tuple regular expressions. In addition, we introduced the notion of 'ellipsis', which indicates the (optional) omission from left-context lexical expressions of tuples; this accounts for spreading.

Two-level implementations either work directly on rules or compile rules into FSTs. For the latter case, we propose an auxiliary finite-state transducer into which multi-tape two-level rules can be compiled. The machine scans 'tuple pairs' instead of pairs of symbols.

The outline of the paper is as follows: Section 2 introduces the root-and-pattern nature of Semitic morphology. Section 3 provides a review of the previous proposals for handling Semitic morphology. Section 4 presents our proposal, extending two-level morphology and proposing a formalism which is adequate for writing non-linear grammars using high level notation. Section 5 applies our model on the Arabic verb. Section 6 presents an auxiliary automaton into which multi-tape two-level rules can be compiled. Finally, section 7 gives concluding remarks.

## 2 ROOT-AND-PATTERN MORPHOLOGY

Non-linear **root-and-pattern** morphology is best illustrated in Semitic. A Semitic stem consists of a **root** and a **vowel melody**, arranged according to a **canonical pattern**. For example, Arabic /*kuttib*/ 'caused to write' is composed from the root morpheme {ktb} 'notion of writing' and the vowel melody morpheme {ui} 'perfect passive'; the two are arranged according to the pattern morpheme {CVCCVC} 'causative'.

Table 1 (next page) gives the Arabic perfective verbal forms (from [McCarthy 1981]).[1]

---

*Supported by a Benefactor Studentship from St John's College. This research was done under the supervision of Dr Stephen G. Pulman whom I thank for guidance, support and feedback. Thanks to Dr John Carroll for editorial comments, Arturo Trujillo for useful 'chats' and Tanya Bowden for Prolog tips.

[1] As indicated by [McCarthy 1981], the data in Table 1 provides stems in underlying morphological forms. Hence, it should be noted that: mood, case, gender and number marking is not shown; many stems experience phonological processing to give surface forms, e.g. /*nkatab*/ → /*ʔinkatab*/ (form 7); the root morphemes shown are not cited in the literature in all forms, e.g. there is no such verb as */*takattab*/ (form 5), but there is /*takassab*/ from the root morpheme {ksb}; the quality of the second vowel in form I is different from one root to another, e.g. /*qatal*/ 'to kill', /*qabil*/ 'to accept', /*kabur*/ 'to become big', from the root morphemes {qtl}, {qbl} and {kbr}, respectively. Some forms do not occur in the passive.

## Table 1 Arabic Verbal Stems

|    | Active    | Passive   |     | Active    | Passive   |
|----|-----------|-----------|-----|-----------|-----------|
| 1  | katab     | kutib     | 11  | ktaabab   |           |
| 2  | kattab    | kuttib    | 12  | ktawtab   |           |
| 3  | kaatab    | kuutib    | 13  | ktawwab   |           |
| 4  | ʔaktab    | ʔuktib    | 14  | ktanbab   |           |
| 5  | takattab  | tukuttib  | 15  | ktanbay   |           |
| 6  | takaatab  | tukuutib  | Q1  | daħraj    | duħrij    |
| 7  | nkatab    | nkutib    | Q2  | tadaħraj  | tuduħrij  |
| 8  | ktatab    | ktutib    | Q3  | dħanraj   | dħunrij   |
| 9  | ktabab    |           | Q4  | dħarjaj   | dħurjij   |
| 10 | staktab   | stuktib   |     |           |           |

Moving horizontally across the table, one notices a change in vowel melody (active {a}, passive {ui}); everything else remains invariant. Moving vertically, a change in canonical pattern occurs; everything else remains invariant.

[Harris 1941] suggested that Semitic stem morphemes are classified into: **root morphemes** consisting of consonants and **pattern morphemes** consisting of vowels and affixes. Morphemes which fall out of the domain of the root-and-pattern system, such as particles and prepositions, are classified as belonging to a third class consisting of successions of consonants and vowels. The analysis of /kuttib/ produces: the root {ktb} 'notion of writing' and the pattern {_u_:i_} 'causative - perfect passive' (where _ indicates a consonant slot, and : indicates gemination).

[McCarthy 1981] provided a deeper analysis under the framework of autosegmental phonology [Goldsmith 1976]. Here, morphemes are classified into: **root morphemes** consisting of consonants, **vocalism morphemes** consisting of vowels, and **pattern morphemes** which are CV-skeleta.[2] Each sits on a separate tier in the autosegmental model, and they are coordinated with association lines according to the principles of autosegmental phonology; when universal principles fail, language specific rules apply. The analysis of /kuttib/ produces three morphemes, linked as illustrated below.

**Fig. 1** Autosegmental analysis of /kuttib/

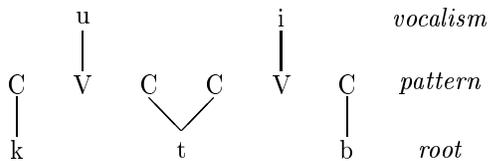

Similarly, one can describe nominals such as /kitaab/ 'book', /kutub/ 'books', /kaatib/ 'writer', /kitaaba/ 'writing' and /katiiba/ 'squadron' etc.

---

[2]The analysis of Arabic here is based on CV theory [McCarthy 1981]. Moraic [McCarthy and Prince 1990a] and affixational [McCarthy 1992] analyses will be discussed in a future work.

## 3 COMPUTATIONAL MODELS

In the past decade, two-level morphology, introduced by [Koskenniemi 1983], has become ubiquitous. In section 3.1, we shall take a brief look at two-level morphology. Section 3.2 gives a brief review of the previous proposals for dealing with Semitic non-linear morphology. Section 3.3 looks at the development of the formalism which we have chosen for our proposal.

### 3.1 Two-Level Morphology

This approach defines two levels of strings in recognition and synthesis: lexical and surface. the former is a representation of lexical strings; the latter is a representation of surface strings. A mapping scheme between the two levels is described by rules which are compiled into FSTs; the set of FSTs run in parallel. One case of two-level rules takes the following form:

$$a : b \quad \Rightarrow \quad c : d \_\_\_ e : f$$

i.e. lexical $a$ corresponds to surface $b$ when preceeded by lexical $c$ corresponding to surface $d$ and followed by lexical $e$ corresponding to surface $f$. The operator is one of four types: $\Rightarrow$ for a context restriction rule, $\Leftarrow$ for a surface coercion rule, $\Leftrightarrow$ for a composite rule (i.e. a composition of $\Rightarrow$ and $\Leftarrow$), and $/\Leftarrow$ for an exclusion rule. Here is an example from [Ritchie 1992]:

**Fig. 2** Two-level description of *moved*

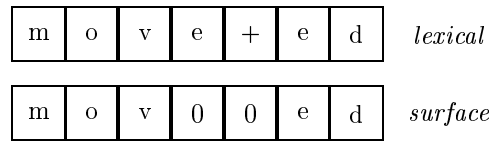

The process can be described by the rules:

$$X : X \quad \Rightarrow \quad \_\_\_ \qquad (1)$$
$$+ : 0 \quad \Rightarrow \quad \_\_\_ \qquad (2)$$
$$e : 0 \quad \Rightarrow \quad v : v \_\_\_ + : 0 \qquad (3)$$

Rule 1 is the **default rule**, where a lexical character appears on the surface. Rule 2 is the **boundary rule**, where the lexical morpheme boundary symbol is deleted on the surface (i.e. surfaces as '0'). Rule 3 states the deletion of lexical [e] in {move} in the context shown.

One can see that two-level morphology is highly influenced by concatenative morphology: the first requirement for a surface form to be related to a lexical form, given by [Ritchie 1992], states that "the lexical tape is the *concatenation* of the lexical forms in question..." (italics mine). This makes it extremely difficult, if not impossible, to apply the autonomous morphemes of Semitic to mainstream two-level notation.

## 3.2 Previous Proposals

Working within standard two-level morphology, [Kataja and Koskenniemi 1988] went around the problem. Nominal forms, such as /kitaab/ 'book', were entered in the lexicon. Verbal forms were derived by a 'lexicon component'. A verb, such as /nkutib/ (form 7), has the lexical entries

$$\Sigma_2^* \; k \; \Sigma_2^* \; t \; \Sigma_2^* \; b \; \Sigma_2^*$$
$$n \; \Sigma_1 \; u \; \Sigma_1 \; i \; \Sigma_1$$

where $\Sigma_1$ is the alphabet of the root and $\Sigma_2$ the alphabet of the vocalism/affixes. The lexicon component takes the intersection of these two expressions and produces /nkutib/. Now /nkutib/ is fed on the lexical tape of a standard two-level system which takes care of conditional phonetic changes (assimilation, deletion, etc.) and produces /ʔinkutib/.[3] A similar approach was used by [Lavie et al. 1988] for Hebrew using a 'pre-lexical compiler'.

[Kay 1987] proposed a finite-state approach using four tapes for root, CV-skeleton, vowel melody and surface, each having an independent head, i.e. the machine can scan from one lexical tape without moving the head on other lexical tapes. The absence of motion is indicated by *ad hoc* notation coded in the lexical strings.

[Beesley 1991], working on Arabic, implemented a two-level system with 'detours', where, according to [Sproat 1992, p. 163-64], detouring involves multiple dictionaries being open at a time, one for roots and one for templates with vowels pre-compiled (as in Harris' description).

Other non two-level models were proposed (there is no place here for a review of these works): [Kornai 1991] proposed a model for autosegmental phonology using FSTs, where non-linear autosegmental representations are coded as linear strings. [Bird and Ellison 1992] proposed a model based on *one*-level phonology using FSA to model representations and rules. [Wiebe 1992] proposed modelling autosegmental phonology using multi-tape FSTs, where autosegmental representations are coded in arrays.

[Pulman and Hepple 1993] proposed a formalism for bidirectional segmental phonological processing, and proposed using it for Arabic. The next subsection presents the development of this formalism.

## 3.3 Previous Formalisms

[Black et al. 1987] pointed out that previous two-level rules (cf. §3.1) affect one character at a time and proposed a formalism which maps between (equal numbered) *sequences* of surface and lexical characters of the form,

$$\text{SURF} \Leftrightarrow \text{LEX}$$

A lexical string maps to a surface string iff they can be partitioned into pairs of lexical-surface subsequences, where each pair is licenced by a rule. [Ruessink 1989] added explicit contexts and allowed unequal sequences. [Pulman and Hepple 1993] developed the formalism further, allowing feature-based representations interpreted via unification.

The developed formalism is based on the existence of only two levels of representation: surface and lexical. Two types of rules are provided:

$$\text{LSC - SURF - RSC} \Rightarrow \text{LLC - LEX - RLC}$$
$$\text{LSC - SURF - RSC} \Leftrightarrow \text{LLC - LEX - RLC}$$

where
| | | |
|---|---|---|
| LSC | = | left surface context |
| SURF | = | surface form |
| RSC | = | right surface context |
| LLC | = | left lexical context |
| LEX | = | lexical form |
| RLC | = | right lexical context |

The special symbol * indicates an empty context, which is always satisfied. The operator $\Rightarrow$ states that LEX *may* surface as SURF in the given context, while the operator $\Leftrightarrow$ adds the condition that when LEX appears in the given context, then the surface description *must* satisfy SURF. The later caters for obligatory rules.

The advantage of this formalism over others is that it allows *inter alia* mappings between lexical and surface strings of unequal lengths.[4]

Rules 1- 3 can be expressed in this formalism as follows:[5]

$$* - X - * \;\; \Rightarrow \;\; * - X - * \quad (4)$$
$$* - \;\; - * \;\; \Rightarrow \;\; * - + - * \quad (5)$$
$$* - \;\; - * \;\; \Leftrightarrow \;\; v - e - + \quad (6)$$

Pulman and Hepple proposed using the formalism for Arabic in the following manner: surface /kuttib/ can be expressed with the rule:

$$* - C_1 u C_2 C_2 i C_3 - * \;\; \Rightarrow \;\; + - C_1 C_2 C_3 - +$$

where $C_n$ represents the *n*th radical of the root. They conclude that their representation is closer to the linguistic analysis of Harris than McCarthy. The only disadvantage is that lexical elements, sc. pattern and vocalism, appear in rules resulting in one rule per template-vocalism.

## 4 A MULTI-TAPE TWO-LEVEL APPROACH

Now we present our proposed model. Section 4.1 defines a multi-tape two-level model. Section 4.2 expands the formalism presented in section 3.3 making it a multi-tape two-level formalism.

---

[3] Initial consonant clusters, CC, take a prosthetic /ʔi/.

[4] This allows two-level grammars to handle CV, moraic and infixational analyses which we shall present in a future work.

[5] 0 in rules 1- 3 is indicated here by blank.

## 4.1 A Multi-Tape Two-Level Model

This work follows [Kay 1987] in using three tapes for the lexical level: **pattern tape** (PT), **root tape** (RT) and **vocalism tape** (VT), and one **surface tape** (ST). In synthesis, the lexical tapes are in read mode and the surface tape is in write mode; in recognition, the opposite state of affairs holds. One of the lexical tapes is called the **primary lexical tape** (PLT) through which all lexical morphemes which fall out of the domain of root-and-pattern morphology are passed (e.g. prefixes, suffixes, particles, prepositions). Since characters in PT correspond to those on ST, PT was chosen as PLT.

There is linguistic support for $n$ lexical tapes mapping to one surface tape. As described by [McCarthy 1986], when a word is uttered, it is pronounced in a linear string of segments (corresponding to the linear ST in this model), i.e. the multi-tier representation is linearised. McCarthy calls this process **tier conflation**.

## 4.2 A Multi-Tape Two-Level Formalism

The Pulman-Hepple/Ruessink/Black *et al.* formalism is adopted here with two extensions. The first extension is that all expressions in the lexical side of the rules (i.e. LLC, LEX and RLC) are $n$-tuple regular expressions of the form:

$$(x_1, x_2, \ldots, x_n)$$

If a regular expression ignores all tapes but PLT, the parentheses can be ignored; hence, $(x)$ is the same as $x$ where $x$ is on PLT. Having $n$-tuple lexical expressions and 1-tuple surface expression corresponds to having $n$-tapes on the lexical level and one on the surface.

The second extension is giving LLC the ability to contain **ellipsis**, ... , which indicates the (optional) omission from LLC of tuples, provided that the tuples to the left of ... are the first to appear on the left of LEX. For example, the LLC expression

$$(a) \cdots (b)$$

matches ab, a$x_1$b, a$x_1 x_2$b, a$x_1 x_2 \ldots$b, where $x_i \neq (a)$.

In standard two-level morphology we talk of **feasible pairs**. Here we talk of **feasible tuple pairs** of the form

$$(x_1, x_2, \ldots, x_n) : (y)$$

For example, Rule 8 (see below) gives rise to four feasible tuple pairs ($C_i$, X, ):(X), $1 \leq i \leq 4$. The set of feasible tuple pairs is determined the same way as the set of feasible pairs in standard two-level grammars.

Now that we have presented our proposal, we are ready to apply it to the Arabic data of Table 1.

## 5 ANALYSIS OF THE ARABIC VERB

Section 5.1 presents the default and boundary rules for Arabic in the two-level formalism. Section 5.2 gives rules which handle vocalised-, non-vocalised-, and partially vocalised texts. Finally, we shall see the use of ellipsis to account for gemination and spreading in section 5.3.

### 5.1 Default and Boundary Rules

The default and boundary rules for Arabic in the multi-tape formalism are:[6]

$$* - X - * \Rightarrow * - X - * \quad (7)$$

$$* - X - * \Rightarrow * - (C, X, \,) - *$$
$$C \in \{c_1, c_2, c_3, c_4\} \quad (8)$$

$$* - X - * \Rightarrow * - (V, \,, X) - *$$
$$V \in \{v_1, v_2\} \quad (9)$$

$$* - \phantom{X} -* \Rightarrow * - + - * \quad (10)$$

$$* - \phantom{X} -* \Rightarrow * - (+, +, +) - * \quad (11)$$

Rule 7 is equivalent to Rule 1. Rule 8 states that any C on the pattern tape and X on the root tape with no transition on the vocalism tape correspond to X on the surface tape. Rule 9 states that any V on the pattern tape and X on vocalism tape with no transition on the root tape correspond to X on the surface tape. Rule 10 is the boundary rule for morphemes which lie out of the domain of root-and-pattern morphology. Rule 11 is the boundary rule for stems.

Here is the derivation of /dḥunrija/ (form Q3) from the three morphemes $\{c_1c_2v_1nc_3v_2c_4\}$,[7] {dḥrj} and {ui}, and the suffix {a} '3rd person' which falls out of the domain of root-and-pattern morphology and, hence, takes its place on PLT.

**Fig. 3a** Form Q3 + {a}

| u | | i | | | + | VT |
|---|---|---|---|---|---|---|
| d | ḥ | | r | j | + | RT |
| $c_1$ | $c_2$ | $v_1$ | n | $c_3$ | $v_2$ | $c_4$ | + | a | + | PT |
| 8 | 8 | 9 | 7 | 8 | 9 | 8 | 11 | 7 | 10 | |
| d | ḥ | u | n | r | i | j | | a | | ST |

The numbers between ST and the lexical tapes indicate the rules which sanction the moves.

We find that default and boundary rules represent a wide range of Semitic stems.

---

[6] Variables are indicated by upper-case letters and atomic elements by lower case-letters.

[7] Note that association lines are indicated implicitly by numbering the CV elements in the pattern morpheme.

## 5.2 Vocalisation

Orthographically, Semitic texts appear in three forms: **consonantal texts** do not incorporate any vowels but *matres lectionis*[8], e.g. *ktb* for /katab/ (form 1, active), /kutib/ (form 1, passive) and /kutub/ 'books', but *kaatb* for /kaatab/ (form 3, active) and /kaatib/ 'writer'; **partially vocalised texts** incorporate some vowels to clarify ambiguity, e.g. *kutb* for /kutib/ (form 1, passive) to distinguish it from /katab/ (form 1, active); and **vocalised texts** incorporate full vocalisation, e.g. *staktab* (form 10, active).

This phenomenon is taken care of by the following rules:

$$* - \quad -* \Rightarrow (X_1) - (V) - (X_2)$$
$$X_1, X_2 \neq \text{vowel} \quad (12)$$

$$* - \quad -* \Rightarrow (P_1, X_1, \ ) - (P, \ , X) - (P_2, X_2, \ )$$
$$P \in \{v_1, v_2\}, \ X = \text{vowel},$$
$$P_1, P_2 \in \{c_1, c_2, c_3, c_4\},$$
$$X_1, X_2 = \text{radical} \quad (13)$$

Rule 12 allows the omission of non-stem vowels (i.e. prefixes and suffixes). Rule 13 allows the omission of stem vowels. Note that the lexical contexts, LLC and RLC, ensure that *matres lectionis* are not omitted in the surface. Here is form Q3 with partial vocalisation on the surface.

**Fig. 3b**  Form Q3 + {a} partially vocalised

| VT | u | | i | | | + | |
|---|---|---|---|---|---|---|---|
| RT | d | ħ | | r | j | | + | |
| PT | $c_1$ | $c_2$ | $v_1$ | n | $c_3$ | $v_2$ | $c_4$ | + | a | + |
| | 8 | 8 | 9 | 7 | 8 | 13 | 8 | 11 | 12 | 10 |
| ST | d | ħ | u | n | r | | j | | | |

One additional rule is required to allow the omission of vowels which experience spreading (see Rule 17 below).

## 5.3 Gemination and Spreading

The only two phonological changes in the Arabic stem are gemination and spreading, e.g. /tukuttib/ (form 5) from the morphemes {$tv_1c_1v_1c_2c_2v_2c_3$}, {ktb} and {ui}. The gemination of the second radical [t] and the spreading of the first vowel [u] can be expressed by Rule 14 and Rule 15, respectively:

$$* - X - * \Rightarrow (c_2, X, \ ) - c_2 - * \quad (14)$$
$$* - X - * \Rightarrow (v_1, \ , X) \cdots - v_1 - * \quad (15)$$

Note the use of ellipsis to indicate that there are elements separating the two [u]s. Form 5 is illustrated below (without boundary symbols).

**Fig. 4**  Form 5

| VT | u | | | i | | | |
|---|---|---|---|---|---|---|---|
| RT | | k | | t | | b | |
| PT | t | $v_1$ | $c_1$ | $v_1$ | $c_2$ | $c_2$ | $v_2$ | $c_3$ |
| | 7 | 9 | 8 | 15 | 8 | 14 | 9 | 8 |
| ST | t | u | k | u | t | t | i | b |

In fact, gemination can be considered as a case of spreading; Rule 14 becomes,

$$* - X - * \Rightarrow (c_2, X, \ ) \cdots - c_2 - * \quad (16)$$

This allows for /tukuttib/ (form 5) and /ktawtab/ (form 12).

We also need to allow a vowel which originally surfaces by spreading to be omitted in the surface in unvocalised words. This is accomplished by the following rule:

$$* - \quad -* \Rightarrow (v_1, \ , X) \cdots (P_1, X_1, \ ) - v_1 - (P_2, X_2, \ )$$
$$X = \text{vowel},$$
$$P_1, P_2 \in \{c_1, c_2, c_3, c_4\},$$
$$X_1, X_2 = \text{radical} \quad (17)$$

Note that the segments in SURF in the above rules do *not* appear in LEX, rather in LLC. This means that, if rules are to be compiled into automata, the automata have to *remember* the segments from LLC.[9] This leads us on thinking about what sort of automata are needed to describe a multi-tape two-level grammar.

## 6 Compilation into Automata

We define the following automaton into which rules can be compiled:

A **multi-tape $\ell$-register auxiliary finite-state automaton** (AFSA) with *n*-tapes consists of: *n* read tapes and heads, a finite state control, and a read-write storage tape of length $\ell$, where $\ell \leq w$, and *w* is the length of the input strings (cf. APDA in [Hopcroft and Ullman 1979]). The automaton is illustrated in Fig. 5 (next page).[10]

In one move, depending on the state of the finite control, along with the symbols scanned by the input and storage heads, the AFSA may do any or all of the following:

---

[8] 'Mothers of reading', these are consonantal letters which play the role of vowels, and are represented in the pattern morpheme by VV (e.g. /aa/, /uu/, /ii/). *Matres lectionis* cannot be omitted from the orthographic string.

[9] If the implementation works directly on rules, this can be achieved by unification.

[10] $\ell = \lambda$ in the diagram.

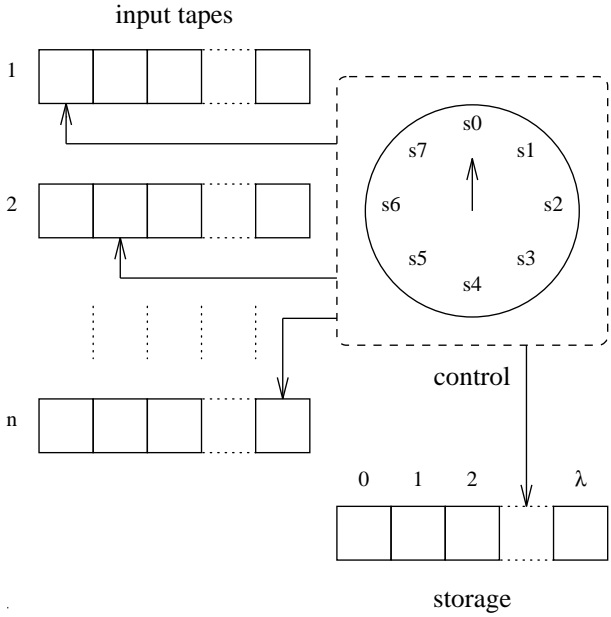

Fig. 5 AFST

- change state;
- move its $n$ input heads independently one position to the right;
- print a symbol on the cell scanned by the storage head and (optionally) move that head one position to the right or left.

More formally an AFSA is a sextuple of the form $(Q, \Sigma, \Gamma, \delta, q_0, F)$, where:

- $Q$ is a finite set of states;
- $\Sigma$ is the machine's alphabet;
- $\Gamma \subseteq \Sigma$ is the storage alphabet;
- $\delta$ is the transition function, a map from $Q \times \sigma \times \Gamma$ to $Q \times \Gamma \times \{L, R\}$, where $\sigma$ is $(\sigma_1, ..., \sigma_n)$ and $\sigma_i \in \Sigma$;
- $q_0 \in Q$ is the initial state;
- $F \subseteq Q$ is the set of final states.

The transition function $\delta(p, \sigma, r) = (q, w, m)$ iff the machine can move from state $p$ to state $q$ while scanning the $n$-tuple $\sigma$ from the input tapes and $r$ from the current storage cell, and upon entering state $q$, writes the symbol $w$ onto the current storage cell and moves the storage head according to $m \in \{L, R\}$.

A **multi-tape $\ell$-register auxiliary finite-state transducer** (AFST) with $n$ input tapes and $k$ output tapes is an AFSA with $(n + k)$-tapes. AFSTs behave like AFSAs, but scan tuple pairs.

Note that an AFST with $n = k = 1$ and $\ell = 0$ is equivalent to a FST.

The rules are compiled into AFSTs in the same lines of standard two-level morphology. We shall use a special case of AFSTs: We hypothesise that, in lines with tier conflation, for all morphological processes, $k$=1

(i.e. one surface tape); further, we assume that, unless one proves otherwise, all morphological processes require that $\ell \leq 1$ (hence, we shall ignore $m$ in $\delta$).

For Semitic, $n$=3. The AFST for Rule 15 is illustrated below.

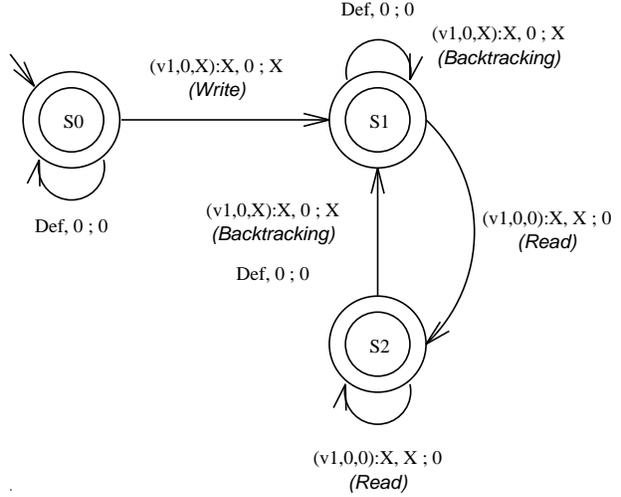

Fig. 6 AFST for Rule 15

Transitions marked with Def (for default) take place when $\sigma$ is a feasible tuple pair, other than those explicitly shown. The empty string is represented by 0. The transitions are:

- $\delta(s_0, Def, 0) = (s_0, 0)$ allows strings not related to this rule to be accepted;
- $\delta(s_0, (v_1, 0, X) : X, 0) = (s_1, X)$ enters the rule writing X in the storage cell;
- $\delta(s_1, (v_1, 0, X) : X, 0) = (s_1, X)$ and $\delta(s_2, (v_1, 0, X) : X, 0) = (s_1, X)$ ensure backtracking;
- $\delta(s_1, Def, 0) = (s_1, 0)$ represents ellipsis;
- $\delta(s_1, (v_1, 0, 0) : X, X) = (s_2, 0)$ retrieves the contents of the storage cell;
- $\delta(s_2, (v_1, 0, 0) : X, X) = (s_2, 0)$ allows consecutive reading operations, e.g. [aa] in /takaatab/ (form 6).
- $\delta(s_2, Def, 0) = (s_1, 0)$ allows non-consecutive reading operations, e.g. the three [a]s in /takattab/ (form 5).

## 7 CONCLUSION

This paper has shown that a multi-tape two-level approach using the Pulman-Hepple/Ruessink/Black *et al.* formalism with the extensions mentioned is capable of describing the whole range of Arabic stems.

Why do we need storage in the automata? It is known that an automaton with finite storage can be replaced with a larger one without storage (a simple solution is to duplicate the machine for each case); hence, using finite storage (especially with $\ell \leq 1$ and a small finite set of $\Gamma$) does not give the machine extra power. The reason for using storage is to minimise the number of machines and states.

With regards to the implementation, first we implemented a small system in order to test the usage of AFSTs in our model. Once this was established, we made a second implementation based on the work of [Pulman and Hepple 1993]. This implementation differs from theirs as follows: Lexical expressions are $n$-tuples, i.e. implemented as lists-of-lists instead of lists-of-characters. A facility to check ellipsis in rules was added. The lexicon consists of multiple trees, one tree per tape. Finally, a morphosyntactic parser was added.

We conclude this paper by looking at the possibility of using our model for tonal languages.

## 7.1 Beyond Semitic

This approach *may* be capable of describing other types of non-linear morphology, though we have not yet looked at a whole range of examples. The following may form a theoretical framework for a number of non-linear phenomena.

Consider suprasegmental morphology in tonal languages. Tense in Ngbaka, a language of Zaire, is indicated by tone, e.g. {kpolo} 'return' gives /kpòlò/ (Low), /kpōlō/ (Mid), /kpòló/ (Low-High), and /kpóló/ (High) [Nida 1949]. This can be expressed with the stem morpheme {kpolo} on one tape and the tonal morphemes {L}, {M}, {LH} and {H} on a second tape with the following rules:

$$* - C - * \Rightarrow * - C - * \quad (18)$$
$$* - V - * \Rightarrow * - V - * \quad (19)$$
$$* - T - * \Leftrightarrow (V,) - (,T) - * \quad (20)$$

where C is a consonant, V is a vowel and T is a tonal segment (these rules are for the above data only). The transitions for /kpòló/ are shown below:

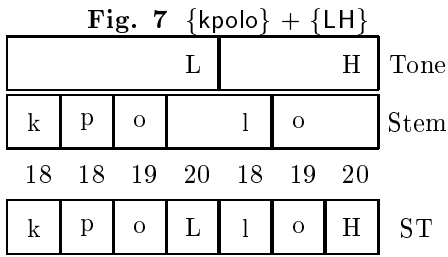

Fig. 7 {kpolo} + {LH}

For all other cases one needs to add a rule for spreading the tonal morpheme.

## 7.2 Future Work

Currently, we are looking at describing the Semitic stem using moraic [McCarthy and Prince 1990a] and affixational [McCarthy 1992] analyses of Semitic stems.

Another area of interest is to look at the formal properties of the formalism and of the AFSM.

# References

[Beesley 1991] K. Beesley. 'Computer Analysis of Arabic Morphology.' B. Comrie and M. Eid (eds.) *Perspectives on Arabic Linguistics III*.

[Bird and Ellison 1992] S. Bird and T. Ellison. *One Level Phonology*. Edinburgh research Papers in Cognitive Science, No. EUCCS/RP-51 (updated version 1993).

[Black et al. 1987] A. Black, G. Ritchie, S. Pulman, G. Russel. 'Formalisms for Morphographemic Description.' *EACL-3*.

[Goldsmith 1976] J. Goldsmith. *Autosegmental Phonology*, Doctoral dissertation, MIT. Published later as *Autosegmental and Metrical Phonology* (Oxford 1990.)

[Harris 1941] Z. Harris. 'Linguistic Structure of Hebrew.' *Journal of the American Oriental Society*: 61.

[Hopcroft and Ullman 1979] J. Hopcroft and J. Ullman. *Introduction to Automata Theory, Languages, and Computation*. (Addison-Wesley).

[Kataja and Koskenniemi 1988] L. Kataja and K. Koskenniemi. 'Finite State Description of Semitic Morphology.' *COLING-88*.

[Kay 1987] M. Kay. 'Nonconcatenative Finite-State Morphology.' *ACL Proceedings, 3rd European Meeting*.

[Kornai 1991] A. Kornai. *Formal Phonology*. Ph.D. thesis, Stanford University.

[Koskenniemi 1983] K. Koskenniemi. *Two Level Morphology*. Ph.D. thesis, University of Helsinki.

[Lavie et al. 1988] A. Lavie, A. Itai, U. Ornan. 'On the Applicability of Two Level Morphology to the Inflection of Hebrew Verbs.' *Proceedings of ALLC III*.

[McCarthy 1981] J. J. McCarthy. 'A Prosodic Theory of Nonconcatenative Morphology.' *LI* 12.

[McCarthy 1986] J. J. McCarthy. 'OCP effects: gemination and antigemination' *LI* 17.

[McCarthy and Prince 1990a] J. J. McCarthy and A. S. Prince. 'Prosodic Morphology and Templatic Morphology.' In M. Eid and J. McCarthy (eds.) *Perspectives on Arabic Linguistics II*.

[McCarthy 1992] J. J. McCarthy. 'Template Form in Prosodic Morphology.' (to appear in the proceedings of the Formal Linguistics Society of Mid-America III.

[Nida 1949] E. Nida. *Morphology: The Descriptive Analysis of Words*. (University of Michigan Press.)

[Pulman and Hepple 1993] S. Pulman and M. Hepple. 'A feature-based formalism for two-level phonology: a description and implementation.' *Computer Speech and Language* 7.

[Ritchie 1992] G. Ritchie. 'Languages Generated by Two-Level Morphological Rules.' *CL* 18.

[Ruessink 1989] H. Ruessink. 'Two Level Formalism.' *Utrecht Working Papers in NLP*, No. 5.

[Sproat 1992] R. Sproat. *Morphology and Computation*. (Cambridge, Mass.: MIT)

[Wiebe 1992] B. Wiebe. *Modelling Autosegmental Phonology with Multi-Tape Finite State Transducers*. M.Sc. Thesis. Simon Fraser University.